\documentclass[runningheads]{llncs}

\usepackage{eccv}

\usepackage{eccvabbrv}

\usepackage{graphicx}
\usepackage{booktabs}
\usepackage{algorithm}
\usepackage{algorithmic}
\usepackage{booktabs}
\usepackage{amsmath, amssymb, centernot}
\usepackage{multirow} 
\usepackage{appendix}
\usepackage[numbers]{natbib}
\usepackage{graphicx}   
\usepackage{mathtools}  
\usepackage{stackengine}
\usepackage{tikz}
\usepackage{pgfplots}
\pgfplotsset{compat=1.18}
\usepackage[accsupp]{axessibility}  

\usepackage[pagebackref,breaklinks,colorlinks,citecolor=eccvblue,hypertexnames=false]{hyperref}

\usepackage{orcidlink}
\definecolor{myblue}{RGB}{0,0,255}

\begin{document}

\title{SPARC: Scalable Path-Specific Counterfactual Fairness via Causal Conditional Independence}

\titlerunning{SPARC: Scalable Path-Specific Counterfactual Fairness}

\author{Bowei Tian\inst{1}\orcidlink{0009-0005-7275-7955} \and
Yexiao He\inst{1}\orcidlink{0000-0002-4675-7733} \and
Ziyao Wang\inst{1}\orcidlink{0000-0002-1394-9587} \and
Meng Liu\inst{1}\orcidlink{0009-0000-4994-1440} \and
Yongkai Wu\inst{2}\orcidlink{0000-0002-7313-9439} \and
Ang Li\inst{1}\orcidlink{0000-0002-4990-1729}
}

\authorrunning{B.~Tian et al.}

\institute{
University of Maryland, College Park, MD 20742, USA\\
\email{\{btian1,yexiaohe,ziyaow,mengliu,angliece\}@umd.edu}
\and
Clemson University, Clemson, SC 29634, USA\\
\email{yongkaw@clemson.edu}
}

\maketitle

\begin{abstract}
Deep learning models exhibit fairness concerns when predictions are inadvertently influenced by sensitive attributes. However, existing attempts to make Path-Specific Counterfactual Fairness optimizable rely on estimating marginal potential outcome probabilities—an approach that fundamentally requires high-dimensional conditional density estimation and breaks down in modalities such as medical images, where the curse of dimensionality renders reliable estimation infeasible. To address this limitation, we reduce the problem of enforcing Path-Specific Counterfactual Fairness to a causal conditional independence constraint and prove that satisfying this constraint is sufficient to eliminate the unfair causal effect. This reduction replaces intractable counterfactual estimation with a discriminative optimization objective that remains scalable in high-dimensional settings.
\end{abstract}

\section{Introduction}
\label{sec:intro}

Deep learning models have achieved impressive success, often reaching or exceeding human expert-level performance across various tasks \citep{litjens2017survey, lee2017deep, rajpurkar2018deep}. While these advanced models have improved the accuracy and efficiency of various tasks, they have also raised concerns about the fairness and reliability of AI-driven decisions. It has been revealed that many deep learning models unintentionally incorporate sensitive demographic attributes (e.g., race, gender, age) into their decision-making process, potentially leading to biased predictions and compromised equality \citep{doi:10.1126/science.aax2342, rajkomar2018fairness}. For example, Obermeyer et al. \citep{doi:10.1126/science.aax2342} revealed that some algorithms commercially deployed by major U.S. healthcare systems (e.g., Optum’s Impact Pro) systematically assigned lower risk scores to Black patients than to equally sick White patients, as it learned to associate health status with healthcare costs rather than actual medical needs. This systematic bias undermines the trustworthiness of AI systems and raises ethical concerns regarding their deployment in real-world settings \citep{Gichoya_2022, doi:10.1073/pnas.1919012117, seyyedkalantari2020chexclusionfairnessgapsdeep}.  
Although sensitive information might be omitted, models often infer these demographics from correlated image patterns, perpetuating bias \citep{chen2024fairnessimprovementmultipleprotected}. Moreover, removing sensitive features could compromise diagnostic precision, as these features often contribute to critical diagnostic signals \citep{zhang2018mitigating, madras2018learningadversariallyfairtransferable, 10.1007/978-3-030-11009-3_34}. Therefore, disentangling the bias-induced and task-relevant contributions of these sensitive attributes is of great interest.

Previous studies on mitigating bias in deep learning applications can be broadly categorized into data-centric and algorithmic approaches. 
Data-centric approaches focus on improving the diversity and balance of datasets to reduce bias. For instance, Burlina et al. \citep{burlina2020addressingartificialintelligencebias} employed generative methods for data augmentation, minimizing diagnostic disparities in retinal image classification between light-skinned and dark-skinned individuals. Other approaches \citep{paul2021taratrainingrepresentationalteration,10.1145/3442188.3445879} such as resampling strategies, have been employed to address dataset imbalance issues, ensuring fairer representation of underrepresented subgroups. These approaches seek to address bias from its origin in the data itself, enhancing fairness before any modeling takes place and forming an initial step toward responsible development. However, they do not explicitly model the effect of sensitive attributes and predictions, and cannot ensure that the augmented samples preserve the correct causal relationship between features and labels.

Algorithmic methods address fairness during the model training process by modifying the training objective or employing specialized architectures to reduce performance disparities. For example, Paul et al. \citep{paul2021taratrainingrepresentationalteration} proposed the Training and Representation Alteration (TARA) framework, which adapts domain generalization techniques to improve fairness across demographic groups. Similarly, Zhou et al. \citep{zhou2021radfusionbenchmarkingperformancefairness} developed multimodal methods for pulmonary embolism detection, demonstrating the impact of architectural choices on fairness outcomes. Group-DRO \citep{Sagawa2019DistributionallyRN} aims to improve model fairness and robustness by optimizing worst-group performance, ensuring consistent accuracy across demographic or domain subgroups. Another important debiasing technique is adversarial learning, such as adversarial debiasing (AD) \citep{zhang2018mitigating} and fairness-aware adversarial perturbation (FAAP) \citep{wang2022fairness}. AD uses adversarial learning to mitigate biases within the model by designing the training process as an adversarial game, where a secondary network attempts to detect protected attributes (e.g., gender, race) from the primary model’s output. FAAP takes a distinct approach by generating adversarial perturbations that directly incorporate fairness constraints, evaluating the model’s robustness under fairness-aware noise.

Recent research applies causal inference \citep{pearl2009causality, 
kusner2017counterfactual, chiappa2018pathspecific, chikahara2021learning} to fairness by 
examining how sensitive attributes causally influence model features 
and decisions. One prevalent concept is Counterfactual Fairness 
\citep{kusner2017counterfactual}, which requires that a model's 
prediction for an individual remains unchanged under counterfactual 
variations of the sensitive attribute—effectively removing all its 
causal influence on the decision. Although principled, this 
formulation can be overly restrictive, since some causal pathways 
from the sensitive attribute to the outcome may reflect legitimate, 
domain-relevant information \citep{yang2024limits, jones2024causal}. 
To address this limitation, Chiappa et al. \citep{chiappa2018pathspecific} introduced 
Path-Specific Counterfactual Fairness (PSCF), which distinguishes 
between ``fair'' and ``unfair'' causal paths and removes only the 
latter's effects, preserving meaningful contributions while eliminating 
biased influence. Recent extensions of counterfactual fairness have 
further explored the inherent trade-off between fairness and predictive 
performance \citep{zhou2024counterfactual, tian2024towards}, as well as model-agnostic 
pre-processing approaches based on Structural Causal Models (SCM) 
\citep{chen2025counterfactual}. These causal approaches represent an important step toward 
interpretable and principled fairness, yet existing PSCF methods have been validated exclusively on structured tabular data, and their reliance on marginal potential outcome probabilities renders them fundamentally inapplicable to high-dimensional modalities such as medical images \citep{xu2024addressing, altman2018curse, rathore2026causal,chikahara2021learning}.

To tackle these challenges, we introduce SPARC, the Scalable Path-Specific Counterfactual Fairness to clearly separate bias and contributions originating from 
sensitive attributes in high-dimensional data. Theoretically grounded in causal inference \citep{pearl2009causal, 
kusner2017counterfactual, chiappa2018pathspecific}, the method enables disentanglement of the direct causal influence of sensitive attributes on model predictions across high-dimensional modalities. The code is available at \href{https://github.com/CASE-Lab-UMD/SPARC}{https://github.com/CASE-Lab-UMD/SPARC}. Our contributions can be summarized as follows: 

\begin{itemize}
    \item[$\bullet$] We reformulate PSCF as a scalable optimization process by reducing the counterfactual estimation to a causal conditional independence constraint.
    \item[$\bullet$] We provide theoretical proof that satisfying this constraint is sufficient to achieve PSCF; furthermore, we prove that this optimization process is directly observed in standard fairness metrics.
    \item[$\bullet$] We demonstrate the effectiveness of our approach. Extensive experiments show that our method outperforms baselines, achieving a better trade-off between predictive performance and fairness in high-dimensional modalities.
\end{itemize}

\section{Preliminaries}
\subsection{Causal Effect}
\begin{figure}[tbp]
\centering
    \hspace{-3mm}
    \includegraphics[width=0.8\textwidth]{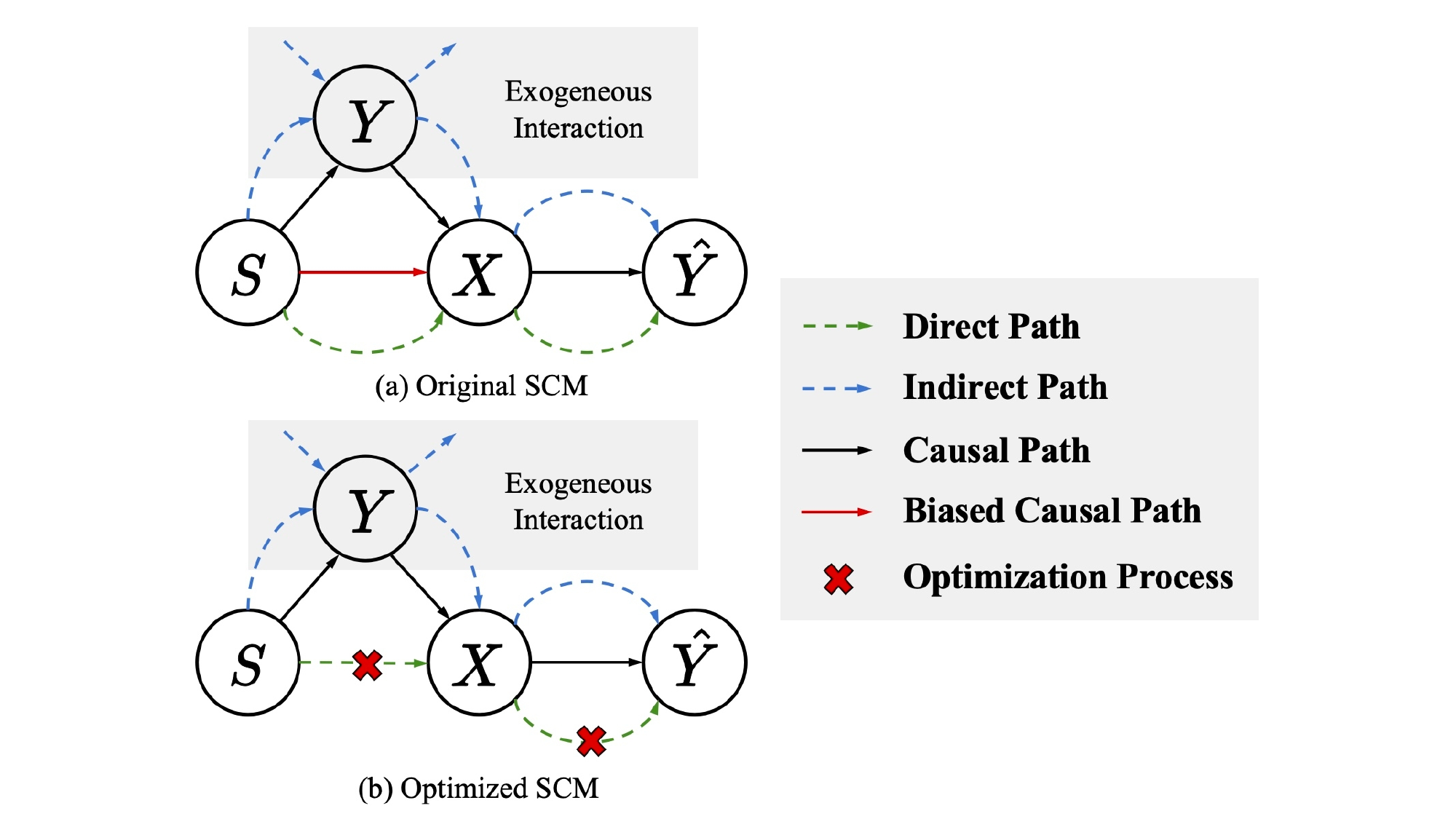} 
    \caption{The theoretical modulation of SCM. Here, $S$ denotes the sensitive attribute, $Y$ represents the ground truth, $\hat{Y}$ is the predicted outcome, and $X$ is the input modality (e.g., X-ray images, text sequence for diagnosis or tabular data), the exogenous interaction refers to the influence of external, unobserved factors that simultaneously affect variables within the model.}
    \label{fig:fig1}
\end{figure}

Causal effect \citep{pearl2000models, zhang2016causal, pearl2022direct} refers to the causal influence that one variable (the cause variable) exerts on another (the outcome variable). Unlike statistical correlation, the causal effect signifies that changes in the cause variable lead to changes in the outcome variable through a direct causal mechanism. We now demonstrate the causal effects within the SCM for the general causal framework, as illustrated in Fig. \ref{fig:fig1}. It is worth noting that if additional information about the exogenous interactions were available (e.g., variations in physicians’ decision-making across regions), the proposed optimization process could be extended to a more sophisticated SCM. In this work, we simplify the problem by not explicitly modeling the exogenous variables. Nonetheless, incorporating such variables would enable a more precise estimation of the causal effects.

\textbf{Definition 1 (Total Causal Effect \citep{pearl2000models})}. The Total Causal Effect (TCE) of a change in the value of sensitive attribute $S$ from $s^-$ to $s^+$ on $\hat{Y}$ is given by:
\begin{equation}
\label{eq:1}
\text{TCE}(S) = P(\hat{Y}_{S \leftarrow s^{+}}) - P(\hat{Y}_{S \leftarrow s^{-}}),
\end{equation}
where the subscript denotes a counterfactual intervention, 
indicating the value $\hat{Y}$ under a hypothetical change of the sensitive attribute \( S \). The TCE measures the influence of $S$ on $\hat{Y}$ as the effect propagates along all causal paths from $S$ to $\hat{Y}$. However, if we consider the influence along only a subset of causal paths from $S$ to $\hat{Y}$, we refer to the resulting effect as the path-specific effect, defined below.

\textbf{Definition 2 (Direct Effect \citep{zhang2016causal, pearl2022direct})}. Given the direct path $\pi_d = \{S\rightarrow X \rightarrow \hat{Y}\}$ and indirect path $\pi_i = \{S \rightarrow Y \rightarrow X \rightarrow \hat{Y}\}$, Direct Effect (DE) represents the path-specific effect of $s^+$ along $\pi_d$, with $s^-$ along $\pi_i$:
\begin{equation}
\label{eq:2}
\text{DE}(S) = 
P(\hat{Y}_{S \leftarrow s^{+}|\pi_d,\, S \leftarrow s^{-}|\pi_i})
- P(\hat{Y}_{S \leftarrow s^{-}}),
\end{equation}
where \( S \leftarrow s^{+}|\pi_d \) denotes a counterfactual intervention 
that sets the sensitive attribute \( S \) to \( s^{+} \) along the direct path \( \pi_d \) , 
and \( S \leftarrow s^{-}|\pi_i \) along the indirect path \( \pi_i \). 
The term \( P(\hat{Y}_{S \leftarrow s^{+}|\pi_d,\, S \leftarrow s^{-}|\pi_i}) \) thus represents a path-specific intervention 
that isolates the direct causal influence of \( S \) on \( \hat{Y} \), 
and the difference with \( P(\hat{Y}_{S \leftarrow s^{-}}) \) 
quantifies the \emph{direct effect} of \( S \) on the model output.

\textbf{Definition 3 (Indirect Effect \citep{zhang2016causal, pearl2022direct})}.
The Indirect Effect (IE) is defined as the difference between the TCE and the DE:
\begin{align}
    \text{IE}(S) = &\, \text{TCE}(S) - \text{DE}(S) \nonumber\\
    = &\, \big(P(\hat{Y}_{S \leftarrow s^{+}}) - P(\hat{Y}_{S \leftarrow s^{-}})\big) \nonumber
     -\,\big(P(\hat{Y}_{S \leftarrow s^{+}|\pi_d,\, S \leftarrow s^{-}|\pi_i}) - P(\hat{Y}_{S \leftarrow s^{-}})\big) \nonumber\\
    = &\, P(\hat{Y}_{S \leftarrow s^{+}}) - P(\hat{Y}_{S \leftarrow s^{+}|\pi_d,\, S \leftarrow s^{-}|\pi_i}),
\end{align}
where IE captures the effect from $S$ that propagated through $\pi_i$.

\label{sec:preliminaries}
\subsection{Path-Specific Counterfactual Fairness}
Fairness ensures predictive models treat individuals equally across sensitive groups such as race, gender, or age \citep{hardt2016equality, dwork2012fairness}. In causal theory \citep{pearl2009causal}, a stronger goal is to achieve Counterfactual Fairness \citep{kusner2017counterfactual}, which can be formulated in an SCM by requiring that predictions remain invariant under counterfactual changes to the sensitive attribute $S$, i.e.,
\begin{align} \label{eq:counterfatual_eq}
P(\hat{Y}_{S \leftarrow s^{+}}) = P(\hat{Y}_{S \leftarrow s^{-}}),
\end{align}
or equivalently, enforcing a zero total causal effect $\text{TCE}(S) = 0$ as defined in Eq.~\ref{eq:1}. This criterion eliminates any causal influence of $S$ on the prediction $\hat{Y}$, ensuring that explicit or implicit biases due to $S$ are removed.

However, enforcing $\text{TCE}(S) = 0$ can be overly restrictive in certain real-world scenarios, where the sensitive attribute $S$ can legitimately influence the ground truth $Y$. For instance, specific diseases may occur predominantly within one gender, making the effect of $S$ on $Y$ meaningful rather than biased. Consequently, a model should allow such legitimate causal contributions through indirect effects while excluding unfair direct influences of $S$ on $\hat{Y}$.

\textbf{Path-Specific Counterfactual Fairness (PSCF)} \citep{chiappa2018pathspecific} provides a principled solution to this issue by distinguishing between \emph{fair} and \emph{unfair} causal paths from $S$ to $\hat{Y}$. Formally, let $\pi_u$ denote the set of unfair causal paths. A model is said to satisfy PSCF if its prediction remains invariant under counterfactual interventions that alter $S$ only along the unfair paths:
\begin{align}
P(\hat{Y}_{S \leftarrow s^{+}|\pi_u}) = P(\hat{Y}_{S \leftarrow s^{-}|\pi_u}).
\end{align}
This ensures that the decision outcome is unaffected by biased causal mechanisms while preserving the legitimate contributions that propagate through fair paths. Since the direct path to the prediction constitutes explicit discrimination, while influences mediated through other variables may capture legitimate information \citep{chiappa2018pathspecific, zhang2018fairness}, the PSCF can be explicitly formulated as:
\begin{align}
P(\hat{Y}_{S \leftarrow s^{+}|\pi_d,\, S \leftarrow s^{-}|\pi_i})
= P(\hat{Y}_{S \leftarrow s^{-}}),
\end{align}
which corresponds to the zero direct effect shown in Eq.~\ref{eq:2}. In the following, we provide detailed practical and theoretical explanations to demonstrate its optimizability.
In finite-sample settings, exact equalities are unattainable; we therefore adopt an $\epsilon$-relaxed formulation, $\epsilon$-PSCF:
\begin{equation}
\big| P(\hat{Y}_{S \leftarrow s^{+}|\pi_d,\, S \leftarrow s^{-}|\pi_i}) - P(\hat{Y}_{S \leftarrow s^{-}}) \big| \leq \epsilon,
\end{equation}
and analogously for the CF criterion in Eq.~\ref{eq:counterfatual_eq}. The empirical fairness scores reported in Tables~\ref{tab:base}--\ref{tab:beta} reflect this difference-based view, where smaller values indicate closer adherence to the corresponding criterion.

\section{SPARC Framework}
\label{sec:method}

In order to eliminate the direct causal effect while preserving the indirect effect through legitimate pathways, we first establish the theoretical prerequisite that guides the entire SPARC framework (Section~\ref{sec:the}), then present two modules that instantiate the theoretical objectives (Section~\ref{sec:pretrain}): a pretraining stage that preserves task-relevant information along the indirect causal path, and an optimization stage that enforces the causal fairness condition.

\subsection{Theoretical Derivability and Observability of SPARC}
\label{sec:the}

In this section, we present the theoretical derivability and observability of the SPARC framework.
Our primary theoretical goal is to achieve PSCF, defined as eliminating the direct causal effect of the sensitive attribute, i.e., $\text{DE}(S)=0$ in Eq. \ref{eq:2}. We establish that the causal conditional independence, $\mathcal{I}(S;X\mid Y) = 0$, where $\mathcal{I}$ denotes Mutual Information (MI) \citep{nalewajski2011elements,shannon1948mathematical}, is sufficient to derive PSCF.

\begin{theorem}[Derivability]\label{theorem:1}
For random variables \( Y \), \( S \), \( X \) and \( \hat{Y} \), the conditional mutual information $\mathcal{I}(S;X \mid Y) = 0$ is a sufficient condition for $\text{DE}(S)=0$.
\end{theorem}

The proofs can be found in Appendix \ref{appendix:the}. This theorem confirms that by successfully training our generator to approach $\mathcal{I}(S;X \mid Y) = 0$, PSCF is satisfied.

While this optimization is theoretically grounded, directly measuring this quantity is challenging in practice. In particular, the variable $X$, e.g., images or prescriptions, is typically high-dimensional. Reliably estimating the joint probability distribution $P(S, X, Y)$ and the conditional probability distribution $P(S, X | Y)$, which are necessary for calculating $\mathcal{I}(S;X \mid Y)$, would require an immense amount of data samples. This refers to the Curse of Dimensionality \citep{altman2018curse, wainwright2019high}, making it practically unfeasible.

We further prove that the causal conditional independence can be directly evaluated by Equalized Odds, bridging the gap between the theoretical objective of SPARC and observable fairness metrics.

\begin{theorem}[Observability]
\label{theorem:2a}
For random variables \( Y \), \( S \), \( X \), and \( \hat{Y} \), $\mathcal{I}(S;X \mid Y) = 0$
is a sufficient condition for Equalized Odds.
\end{theorem}

Theorem \ref{theorem:2a} concludes that achieving the training objective is sufficient to achieve zero equalized odds, thereby paving the way for directly measuring the optimization process of SPARC using observable fairness metrics.

\subsection{Instantiation of SPARC}

\begin{figure}[tbp]
\centering
    \includegraphics[width=0.65\textwidth]{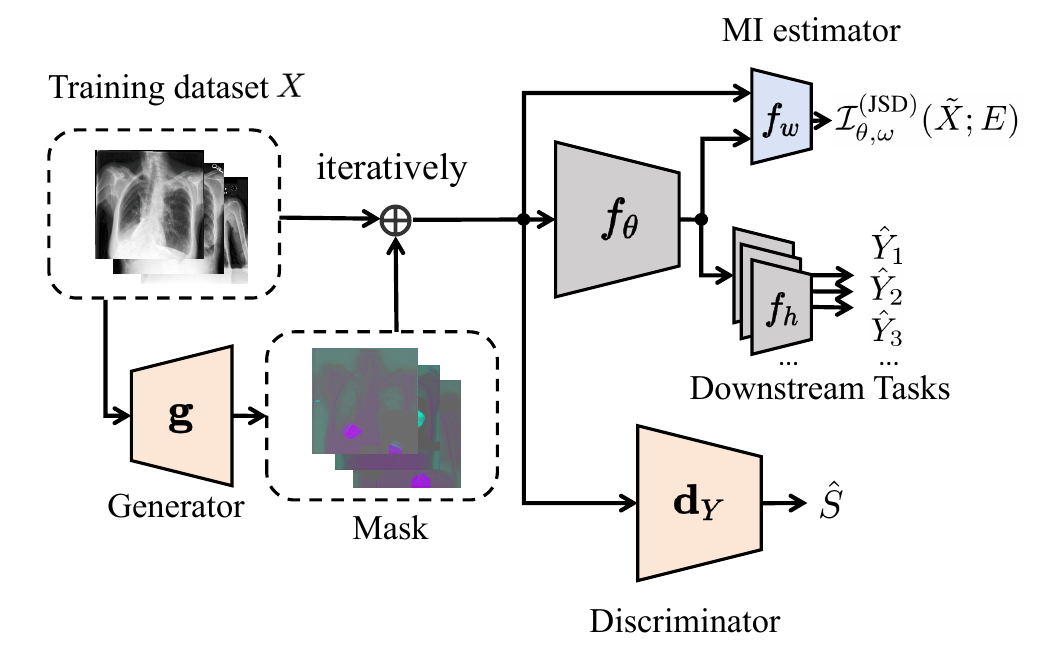}
    \caption{The  instantiation of the SPARC framework. The deployed models, $f_\theta$ and $f_h$ (in grey), are fixed after pretraining. Data utility enhancement (in blue) is applied during pretraining, while adversarial training (in orange) ensures fair masking and maintains original model performance.}
    \label{fig:fig3}
\end{figure}

Figure \ref{fig:fig3} presents an instantiation of the SPARC framework, consisting of the pretraining stage that preserves the indirect causal path and the optimization stage that eliminates the direct path. 
\subsubsection{Pretraining: Preserving the Indirect Causal Path}
\label{sec:pretrain}

Theorem~\ref{theorem:1} requires us to block the direct path $S \to X$ by enforcing $\mathcal{I}(S; X \mid Y) = 0$. However, this intervention must not compromise the indirect path $S \to Y \to \hat{Y}$, which carries legitimate task-relevant information. To ensure that the embedding network retains sufficient information from the masked input $\tilde{X}$ for downstream prediction, we design a data-utility pretraining phase based on Mutual Information. Specifically, we define the embedding as:
\begin{align}
   E = f_\theta(\tilde{X}),
\end{align}
where $\tilde{X} = X \oplus \mathbf{g}(X)$ belongs to the source domain, and the operation $\oplus$ denotes iterative addition with both upper and lower constraints. We employ the Jensen-Shannon MI estimator \cite{hjelm2018learning, nowozin2016f, li2020tiprdc} to maximize the mutual information between $\tilde{X}$ and $E$, which approximates a lower bound on MI:
\begin{align}
&\mathcal{I}(\tilde{X}; E) \geq \mathcal{I}^{(\text{JSD})}_{\theta, \omega}(\tilde{X}; E) \nonumber\\
&:= \mathbb{E}_{\tilde{X}, E} \left[ -\sigma\left(-f_{\omega}(\tilde{X}, E)\right) \right] - \mathbb{E}_{\tilde{X}, E'} \left[ \sigma\left(f_{\omega}(\tilde{X}, E')\right) \right]
\end{align}
where \( f_{\omega} \) is a neural network estimator parameterized by \( \omega \), \( \sigma(z) = \log(1 + e^z) \) is the softplus function, and \( E' \) is a shuffled version of \( E \) acting as the negative embedding. The optimization objective is:
\begin{equation}
    \arg \max_{\theta} \max_{\omega} \mathcal{I}^{(\text{JSD})}_{\theta, \omega}(\tilde{X}; E).
    \label{eq:argmax}
\end{equation}
By maximizing this objective, we ensure that the learned representations $E$ retain essential task-relevant information from $\tilde{X}$ despite the causal intervention, preserving the integrity of the indirect path and enabling effective transfer across diverse downstream tasks via fine-tuning on $\hat{Y}_i = f_h(E)_i$.
\subsubsection{Optimization of Causal Conditional Independence}
\label{sec:opt}

To enforce the causal conditional independence condition in Theorem \ref{theorem:1}, we employ an adversarial framework between a generator $\mathbf{g}$ and a conditioned discriminator $\mathbf{d}_Y$. The generator produces masks that, when applied to the input $X$, yield $\tilde{X} = X \oplus \mathbf{g}(X)$. The discriminator attempts to recover the sensitive attribute from the masked input conditioned on $Y$:
\begin{align}
    \hat{S} = \mathbf{d}_Y(\tilde{X}, Y).
    \label{eq:y}
\end{align}
When the generator succeeds in preventing the discriminator from predicting $S$, the conditional independence $\mathcal{I}(S; X \mid Y) = 0$ is approximately achieved.

The generator loss combines three terms guided by the causal objective:
\begin{equation}
\mathcal{L}_{\mathbf{g}} = \mathcal{I}(\hat{S}; S) - \alpha \mathcal{H}(\hat{S}) - \beta \mathcal{I}(\hat{Y}; Y),
\end{equation}
where $\alpha > 0$ is a relatively small value that controls the regularization of entropy loss, $\beta > 0$ balances the accuracy and fairness trade-off, the first term drives the removal of sensitive information, the entropy term $\mathcal{H}(\hat{S})$ prevents trivial flipping of sensitive attributes, and the last term preserves the deployed model's task accuracy. The discriminator tries to maximize its ability to predict the protected attribute \( \hat{S} \) from the perturbed images, given the prior knowledge of diagnosis $Y$ (shown in Eq. \ref{eq:y}), formulated as:
\begin{equation}
    \mathcal{L}_{\mathbf{d}_Y} = -\mathcal{I}(\hat{S}; S),
\end{equation}
where the objective of $\mathcal{L}_{\mathbf{d}_Y}$ acts as the adversarial of $\mathcal{L}_{\mathbf{g}}$, which showcases how the adversarial training works. 
Consequently, the iterative adversarial training ensures that the generator learns to create fair perturbations (by fooling the discriminator) while preserving the original model performance, as the discriminator tries to distinguish which sensitive group each sample belongs to. The overall min-max objective is:
\begin{equation}
    \arg \max_{\mathbf{g}} \min_{\mathbf{d}_Y} -\mathcal{I}(\hat{S}; S) + \alpha \mathcal{H}(\hat{S}) + \beta\mathcal{I}(\hat{Y};Y),
\end{equation}
where $\mathbf{d}_Y$ and $\mathbf{g}$ are updated alternately, with $\alpha$ and $\beta$ set to $0$ during discriminator updates. Furthermore, as established by Theorem~\ref{theorem:2}, the optimization progress can be directly monitored through Equalized Odds, providing an observable metric that bridges our theoretical objective with practical evaluation. Through this process, we achieve the causal fairness objective of SPARC without requiring explicit counterfactual computation.

\section{Experiments}
\subsection{Experiment Settings}
\subsubsection{Datasets} We evaluate on three open-source datasets:

\noindent
\textbf{MIMIC-CXR} \cite{johnson2019mimic}: The MIMIC-CXR dataset, part of the MIMIC initiative, includes over 370,000 chest X-rays from more than 60,000 patients, labeled with conditions like pneumonia, pleural effusion, and lung lesions. This anonymized dataset is valuable for machine learning research, supporting tasks such as automated disease detection and clinical decision support, especially in critical care environments.

\noindent
\textbf{CheXpert} \cite{irvin2019chexpert}: The CheXpert dataset, created by Stanford University, contains over 224,000 chest X-rays annotated for 14 radiological findings, including lung opacity and cardiomegaly. Known for its high-quality labels obtained through a robust NLP-based pipeline, CheXpert is widely used in AI research for disease classification and anomaly detection in medical imaging.

\noindent
\textbf{TCGA-LUAD} \cite{albertina2016cancer}: This dataset from The Cancer Genome Atlas (TCGA) project focuses on lung adenocarcinoma, providing gene expression, mutation profiles, methylation, and clinical data. It is instrumental for research in lung cancer biomarkers, diagnostics, and targeted therapies, supporting advancements in cancer genomics and precision medicine.

\subsubsection{Fairness Metrics}
We evaluate fairness using the following metrics:

\noindent\textbf{Equalized Opportunity ($\text{EOp}$)} \cite{hardt2016equality} is a notion of nondiscrimination with respect to a specified protected attribute, measuring the disparity in true positive rates across sensitive groups. Specifically, 
\begin{equation}
    \text{EOp} = |P(\hat{Y}=1|S=1,Y=1) - P(\hat{Y}=1|S=0,Y=1)|. 
\end{equation}
\noindent\textbf{Equalized Odds ($\text{EOd}$)} \citep{hardt2016equality} quantifies the independence between the predicted diagnosis $\hat{Y}$ and the sensitive attribute $S$, conditioned on the ground truth $Y$, where we measure in the information theory form \citep{ghassami2018fairness, ni2025kernel, zamani2025information}:
\begin{equation}
    \text{EOd} = \mathcal{I}(S;\hat{Y}\mid Y).
\end{equation}

\noindent\textbf{Demographic Parity (DP)} \cite{dwork2012fairness} measures the disparity in prediction rates between different sensitive groups (e.g., $S=0$ vs. $S=1$). Formally, 
\begin{equation}
    \text{DP} = |P(\hat{Y}=1|S=1) - P(\hat{Y}=1|S=0)|,
\end{equation}
where a smaller DP value indicates fewer disparities between sensitive groups in predictions.
$\text{EOd}$ captures how much uncertainty about $\hat{Y}$ can be reduced by knowing $S$, under the condition of $Y$.

\subsection{Baselines}
We compare with the following baselines: Vanilla \citep{he2016deep}, Adversarial Debiasing (AD) \citep{zhang2018mitigating}, Fairness-Aware Adversarial Perturbation (FAAP) \citep{wang2022fairness}, SUBG \citep{YoubiIdrissi2021SimpleDB} and Group-DRO \citep{Sagawa2019DistributionallyRN}. Vanilla uses traditional gradient descent to optimize; AD introduces an adversarial network that attempts to detect sensitive attributes of the primary model’s output; FAAP generates adversarial perturbations that incorporate fairness constraints; SUBG aims to achieve competitive worst-group-accuracy through simple data balancing; and Group-DRO focuses on distributionally robust neural networks to improve generalization for worst-case scenarios across groups.

\subsection{Main Results}
The main results across the three datasets are presented in Table \ref{tab:base}. Our approach demonstrates a notable improvement in fairness while preserving accuracy. For example, compared with AD, we show a considerable improvement in fairness; compared with FAAP, we reach a better accuracy and fairness; benchmarked against Group-DRO, our method consistently achieves better fairness with a boost in accuracy. Therefore, these observations demonstrate the effectiveness of our method. Additionally, we observe a distinct difference between $\text{EOd}$. For example, in TCGA-LUAD dataset, compared with Vanilla, EOd is lower by 5 times, and EOp and DP are approximately lower by 2 times. The improvement of fairness metrics aligns with the training objective and empirically validates Theorem \ref{theorem:2}. {\color{black}To further verify that improvements in observational metrics correspond to reductions in the path-specific direct effect, we conduct a synthetic-SCM validation where $\text{DE}(S)$ is explicitly computable (Appendix~\ref{app:synthetic}). We attribute the performance gain primarily to the path-specific nature of SPARC. Observational baselines act on $\hat{Y}$ versus $S$ without distinguishing causal pathways, which can lead to either over-correction (suppressing $\pi_i$ and hurting accuracy, as observed in AD/FAAP) or under-correction (leaving a residual direct effect). By targeting the direct path while preserving $\pi_i$ through JSD-MI pretraining (Eq.~\ref{eq:argmax}), SPARC tends to achieve a better joint trade-off between accuracy and fairness. To reduce the influence of confounding factors, all methods share the same backbone, training schedule, and tuning protocol (Appendix~\ref{sec:appendix_section}), suggesting that the improvement is unlikely to stem from finer tuning alone.}

\begin{table*}[h!]
\centering
\caption{Main results on MIMIC-CXR, CheXpert, and TCGA-LUAD datasets. All models are pretrained and fine-tuned on ResNet50. Our baselines are Vanilla \citep{he2016deep}, AD \citep{zhang2018mitigating}, FAAP \citep{wang2022fairness}, SUBG \citep{YoubiIdrissi2021SimpleDB} and Group-DRO \citep{Sagawa2019DistributionallyRN}. The best-averaged score among the methods is \textbf{bolded}, while the second-place averaged score is \underline{underlined}.}
\scalebox{0.93}{
\begin{tabular}{c|c|cc|ccc}
\toprule
\textbf{Dataset} & \textbf{Baselines} & \textbf{$\text{ACC}_{\%}$} ($\uparrow$)& \textbf{$\text{AUC}_{\%}$} ($\uparrow$)& \textbf{$\text{EOp}_{e-2}$} ($\downarrow$)& \textbf{$\text{EOd}_{e-3}$} ($\downarrow$)& \textbf{$\text{DP}_{e-2}$} ($\downarrow$)\\ \hline
\multirow{6}{*}{MIMIC-CXR}& Vanilla & \textbf{85.46} \textsubscript{\textpm 0.49} & \textbf{63.85} \textsubscript{\textpm 0.28} & 5.10 \textsubscript{\textpm 0.13}& 10.59 \textsubscript{\textpm 0.23}& 6.31 \textsubscript{\textpm 0.14}\\
& AD & 83.90 \textsubscript{\textpm 0.44} & 62.01 \textsubscript{\textpm 0.29} & 4.58 \textsubscript{\textpm 0.12}& 7.38 \textsubscript{\textpm 0.16}& 5.56 \textsubscript{\textpm 0.10}\\
& FAAP & 83.07 \textsubscript{\textpm 0.36} & 61.05 \textsubscript{\textpm 0.29} & 4.04 \textsubscript{\textpm 0.12}& 6.07 \textsubscript{\textpm 0.13}& 5.20 \textsubscript{\textpm 0.11}\\
& SUBG& 84.02 \textsubscript{\textpm 0.40} & 61.99 \textsubscript{\textpm 0.30} & 4.79 \textsubscript{\textpm 0.09}& 7.78 \textsubscript{\textpm 0.19}& 5.75 \textsubscript{\textpm 0.13}\\
& Group-DRO& 83.20 \textsubscript{\textpm 0.39} & 61.45 \textsubscript{\textpm 0.27} & \underline{2.90} \textsubscript{\textpm 0.08}& \underline{4.40} \textsubscript{\textpm 0.13}& \underline{3.62} \textsubscript{\textpm 0.09}\\
& Ours & \underline{84.26} \textsubscript{\textpm 0.38} & \underline{62.10} \textsubscript{\textpm 0.30} & \textbf{2.02} \textsubscript{\textpm 0.09}& \textbf{2.25} \textsubscript{\textpm 0.10}& \textbf{3.53} \textsubscript{\textpm 0.08}\\ \hline

\multirow{6}{*}{CheXpert}& Vanilla & \textbf{85.87} \textsubscript{\textpm 0.47} & \textbf{65.93} \textsubscript{\textpm 0.31} & 5.05 \textsubscript{\textpm 0.10}& 11.18 \textsubscript{\textpm 0.26}& 8.86 \textsubscript{\textpm 0.16}\\
& AD & 84.45 \textsubscript{\textpm 0.43} & 63.77 \textsubscript{\textpm 0.29} & 3.43 \textsubscript{\textpm 0.08}& 6.31 \textsubscript{\textpm 0.20}& 7.18 \textsubscript{\textpm 0.13}\\
& FAAP & 83.50 \textsubscript{\textpm 0.34} & 62.55 \textsubscript{\textpm 0.28} & 2.86 \textsubscript{\textpm 0.07}& 4.84 \textsubscript{\textpm 0.17}& 5.40 \textsubscript{\textpm 0.09}\\
& SUBG& 84.95 \textsubscript{\textpm 0.42} & 64.26 \textsubscript{\textpm 0.30} & 3.69 \textsubscript{\textpm 0.08}& 8.35 \textsubscript{\textpm 0.25}& 7.73 \textsubscript{\textpm 0.11}\\
& Group-DRO& 83.55 \textsubscript{\textpm 0.36} & 62.80 \textsubscript{\textpm 0.29} & \underline{1.92} \textsubscript{\textpm 0.07}& \underline{3.07} \textsubscript{\textpm 0.15}& \underline{2.36} \textsubscript{\textpm 0.08}\\
& Ours & \underline{85.80} \textsubscript{\textpm 0.42} & \underline{64.99} \textsubscript{\textpm 0.32} & \textbf{0.98} \textsubscript{\textpm 0.05}& \textbf{1.20} \textsubscript{\textpm 0.08}& \textbf{1.16} \textsubscript{\textpm 0.05}\\ \hline

\multirow{6}{*}{TCGA-LUAD}& Vanilla & \textbf{98.47} \textsubscript{\textpm 0.17} & \textbf{98.58} \textsubscript{\textpm 0.13} & 3.57 \textsubscript{\textpm 0.09}& 7.21 \textsubscript{\textpm 0.15}& 11.94 \textsubscript{\textpm 0.21}\\
& AD & 97.55 \textsubscript{\textpm 0.19} & 97.87 \textsubscript{\textpm 0.15} & 3.20 \textsubscript{\textpm 0.10}& 4.08 \textsubscript{\textpm 0.14}& 9.84 \textsubscript{\textpm 0.16}\\
& FAAP & 97.10 \textsubscript{\textpm 0.22} & 97.64 \textsubscript{\textpm 0.18} & 2.75 \textsubscript{\textpm 0.06}& 3.79 \textsubscript{\textpm 0.08}& 9.46 \textsubscript{\textpm 0.14}\\
& SUBG& 97.50 \textsubscript{\textpm 0.14} & 97.80 \textsubscript{\textpm 0.14} & 3.25 \textsubscript{\textpm 0.09}& 5.05 \textsubscript{\textpm 0.13}& 10.60 \textsubscript{\textpm 0.17}\\
& Group-DRO& 96.65 \textsubscript{\textpm 0.24} & 97.05 \textsubscript{\textpm 0.18} & \underline{1.78} \textsubscript{\textpm 0.08}& \underline{3.12} \textsubscript{\textpm 0.10}& \underline{7.30} \textsubscript{\textpm 0.12}\\
& Ours & \underline{97.85} \textsubscript{\textpm 0.20} & \underline{98.20} \textsubscript{\textpm 0.15} & \textbf{1.39} \textsubscript{\textpm 0.07}& \textbf{1.46} \textsubscript{\textpm 0.08}& \textbf{6.89} \textsubscript{\textpm 0.10}\\ \bottomrule
\end{tabular}}
\label{tab:base}
\end{table*}

\subsection{Data Utility Assessment}
\begin{figure}[tbp]
    \centering
    \includegraphics[width=0.75\textwidth]{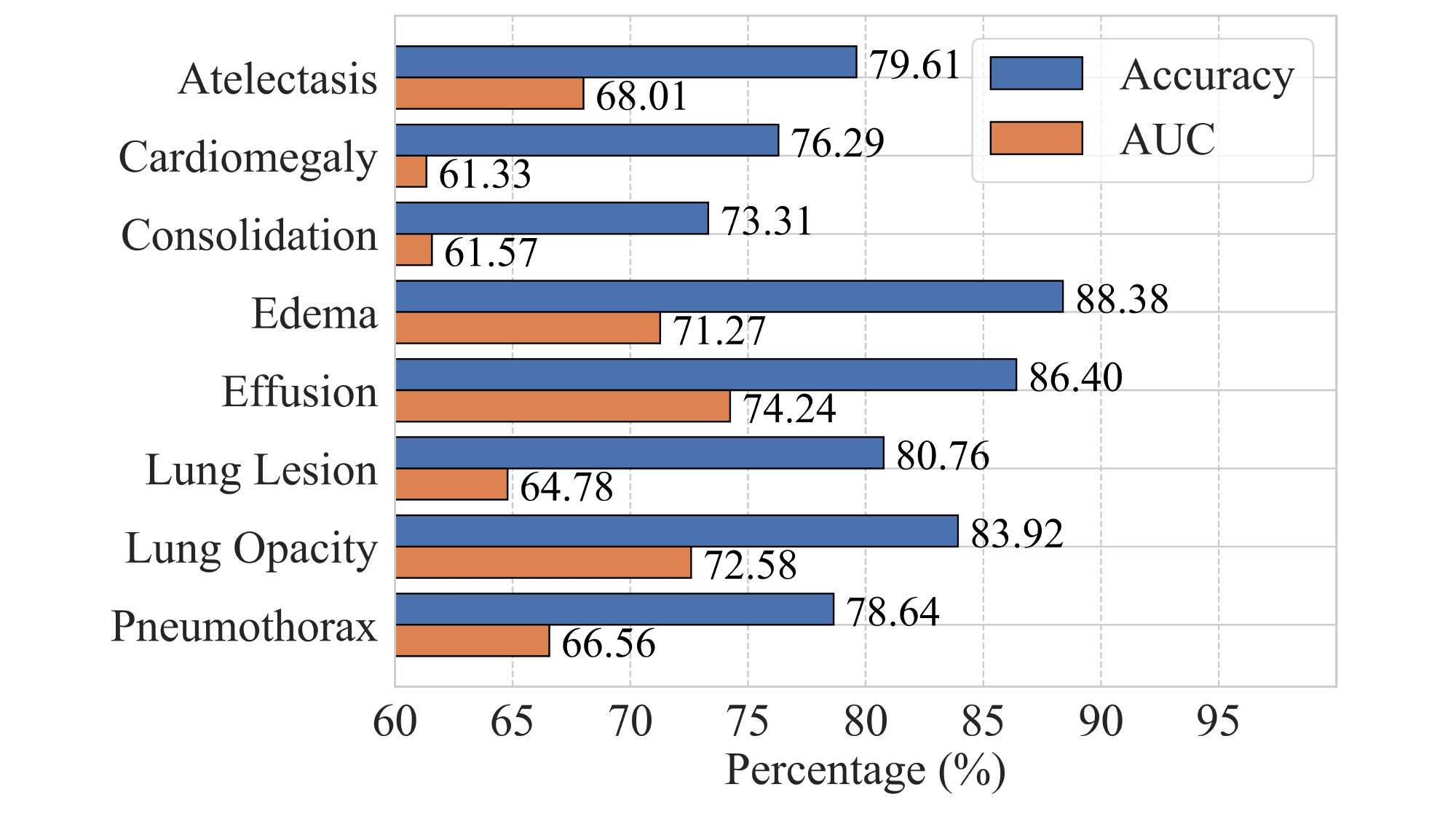} 
    \caption{Data utility evaluations on MIMIC-CXR dataset, showing the fine-tuning performance of \( f_h \) across eight different downstream tasks.}
    \label{fig:fig4}
\end{figure}
We evaluate the data utility of the MIMIC-CXR dataset after applying the MI estimator. Specifically, we start with a pretrained model for ``Pneumonia'' classification, and subsequently fine-tune \( f_h \) across eight different downstream tasks, including the diagnosis of ``Atelectasis'', ``Cardiomegaly'', ``Consolidation'', etc. The experimental results, presented in Fig.~\ref{fig:fig4},  indicate that most tasks achieve an AUC of 65\% and an accuracy of 80\%, demonstrating the robustness and versatility of the MI estimator across downstream tasks.
\begin{table}[t]
\centering
\caption{Ablation studies on different models.}
\scalebox{0.9}{
\begin{tabular}{c|c|cc|ccc}
\toprule
\textbf{Dataset} & \textbf{Models} & \textbf{$\text{ACC}_{\%}$} & \textbf{$\text{AUC}_{\%}$} & \textbf{$\text{EOp}_{e-2}$} & \textbf{$\text{EOd}_{e-3}$} & \textbf{$\text{DP}_{e-2}$} \\ \hline
\multirow{3}{*}{MIMIC}
& Resnet18    & 83.58 & 61.42 & \underline{2.30} & 2.68 & 3.85 \\ 
& Resnet50 & \underline{84.26} & \underline{62.10} & \textbf{2.02} & \underline{2.25} & \underline{3.53} \\ 
& ViT    & \textbf{86.95} & \textbf{63.90} & 3.35 & \textbf{2.07} & \textbf{2.98} \\  \hline
\multirow{3}{*}{CheXpert}
& Resnet18        & 85.02 & 63.85 & \underline{1.22} & \underline{1.55} & \underline{1.48} \\ 
& Resnet50 & \underline{85.80} & \underline{64.99} & \textbf{0.98} & \textbf{1.20} & \textbf{1.16} \\ 
& ViT    & \textbf{86.98} & \textbf{67.53} & 1.78 & 2.51 & 1.76 \\  \hline
\multirow{3}{*}{TCGA}
& Resnet18        & 97.25 & 97.68 & \underline{1.72} & \underline{1.82} & \underline{7.35}\\ 
& Resnet50 & \underline{97.85} & \underline{98.20} & \textbf{1.39} & \textbf{1.46} & \textbf{6.89} \\ 
& ViT    & \textbf{98.45} & \textbf{99.03} & 2.02 & 1.98 & 8.97 \\ \bottomrule
\end{tabular}
}
\label{tab:model}
\end{table}
\subsection{Explainability Analysis}
\begin{figure*}[tbp]
    \centering
    \includegraphics[width=1\textwidth]{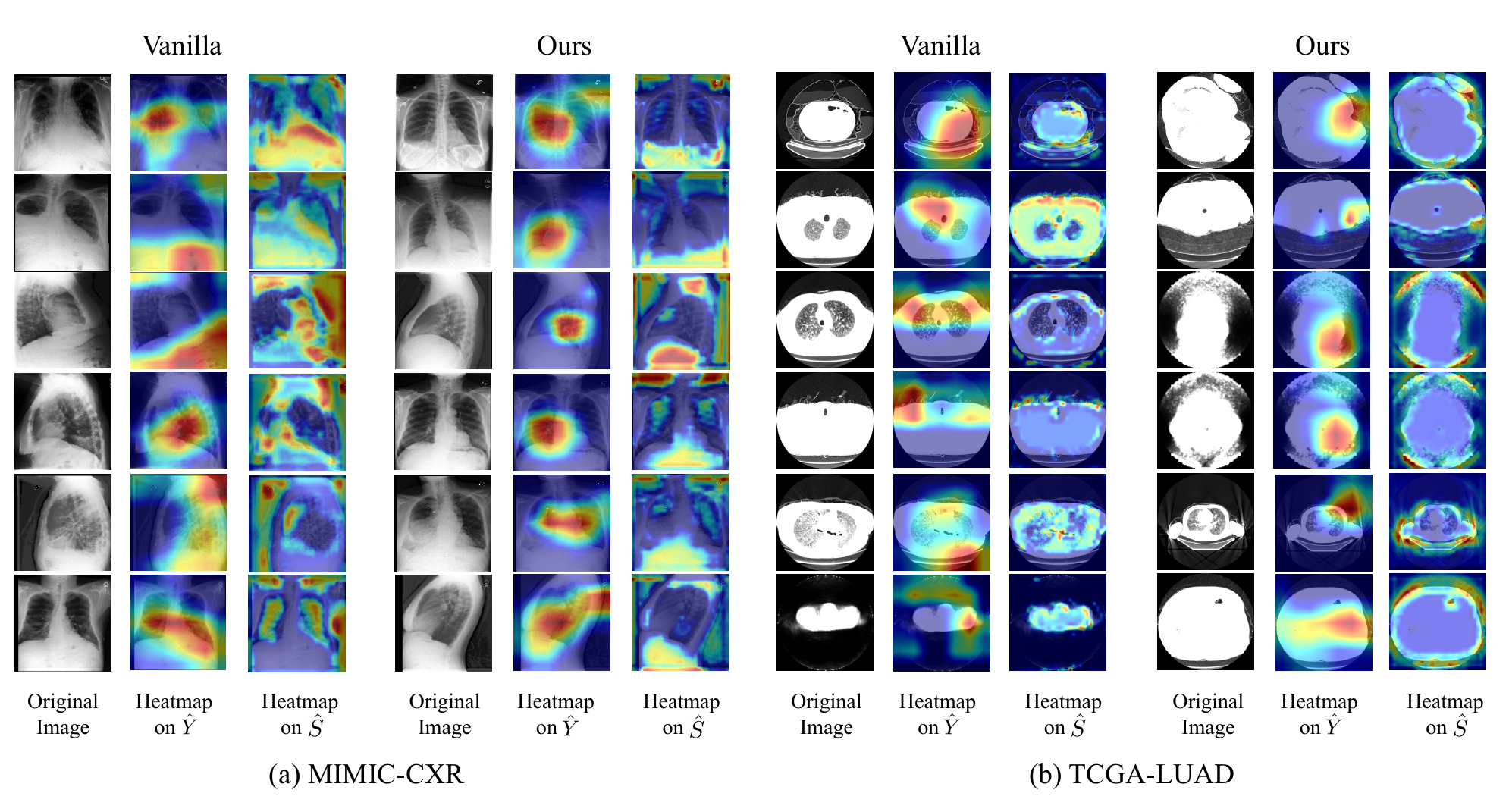} 
    \caption{Explainability analysis on both MIMIC-CXR and TCGA-LUAD datasets.  GradCAM  \citep{selvaraju2017gradcam} is applied on the last convolution layer of $f_\theta$ when applied on $\hat{Y}$, and on $\mathbf{d}_Y$ when applied on $\hat{S}$.}
    \label{fig:wide-image}
\end{figure*}

We perform an explainability analysis on MIMIC-CXR and TCGA-LUAD datasets, where GradCAM \citep{selvaraju2017gradcam} is utilized to generate heatmaps.
These heatmaps visualize the regions that contribute most significantly to the model’s predictions, for both the predicted outcome $\hat{Y}$ and the sensitive attribute prediction $\hat{S}$.
The results in Fig. \ref{fig:wide-image} demonstrate the difference between Vanilla and the SPARC framework. We observe that in Vanilla, the heatmap on $\hat{Y}$ is highly overlapped with that on $\hat{S}$, indicating the bias from sensitive attributes. 
For example, in the MIMIC-CXR dataset, the model may focus on the breast region in female patients, meaning that any variation in “breast shape” could inadvertently affect the diagnosis. This overlap reveals an implicit bias, as the model leverages sensitive attribute information directly in making predictions.
In contrast, our method showcases a clear separation between the regions highlighted for $\hat{Y}$ and  $\hat{S}$ across both datasets. This separation suggests that our method effectively disentangles the information related to the sensitive attribute $\hat{S}$ from the predicted outcome $\hat{Y}$, thereby pruning the direct effect of $S$. By minimizing the overlap, our framework ensures that predictions are driven by medically relevant features rather than sensitive attributes.

\subsection{Demographic Distribution}
\label{subsec:demographic}

We analyze the demographic composition of the three benchmark datasets with respect to gender and age, as shown in Fig.~\ref{fig:gender_distribution} and Fig.~\ref{fig:age_distribution}, respectively. For gender distribution, MIMIC-CXR and TCGA-LUAD exhibit a slight female majority (52.40\% and 53.64\%), while CheXpert is skewed toward male patients (55.48\%). The age distributions reveal more pronounced differences across datasets. MIMIC-CXR is relatively balanced across all four age groups, with each bin comprising roughly 20--27\% of the population. CheXpert shows a similar near-uniform spread but with a slightly higher concentration in the 60--75 range (28.30\%). In contrast, TCGA-LUAD is heavily concentrated in the 60--75 age group (52.88\%), with a negligible proportion of patients under 45 (3.18\%), reflecting the typical age profile of lung adenocarcinoma. These distributional imbalances across both sensitive attributes motivate the need for fairness-aware methods that can generalize across diverse demographic settings.

\subsection{Ablation Studies}
\subsubsection{Model Ablation Study}
We conducted a model ablation study on three different architectures: ResNet18, ResNet50, and Vision Transformers (ViT). The results are presented in Table \ref{tab:model}.
We observe that ViT performs slightly better in terms of accuracy and AUC but with a minor trade-off in fairness metrics. This suggests that our method can enhance fairness across various neural network architectures without significantly compromising performance. For the remaining experiments, we use ResNet50 as it provides a balanced trade-off between performance and fairness.

\subsubsection{Sensitive Attribute Ablation Study} \label{subsec:ablation_attr}
\begin{figure*}[t]
\centering
\begin{minipage}[c]{0.5\textwidth}
\centering
\captionof{table}{Comparison on sensitive attribute: Age ($<60$ vs.\ $\geq 60$).}
\label{tab:age_comparison}
\vspace{9pt}
\resizebox{\textwidth}{!}{%
\begin{tabular}{c|c|cc|ccc}
\hline
\textbf{Dataset} & \textbf{Models} & \textbf{$\text{ACC}_{\%}$} & \textbf{$\text{AUC}_{\%}$} & \textbf{$\text{EOp}_{e-2}$} & \textbf{$\text{EOd}_{e-3}$} & \textbf{$\text{DP}_{e-2}$} \\ \hline
\multirow{2}{*}{MIMIC-CXR}
& Vanilla & \textbf{85.87} & \textbf{64.32} & 4.72 & 9.38 & 5.88 \\
& Ours    & \underline{84.03} & \underline{62.05} & \textbf{1.56} & \textbf{1.45} & \textbf{2.27} \\
\hline
\multirow{2}{*}{CheXpert}
& Vanilla & \textbf{85.95} & \textbf{66.15} & 4.68 & 10.42 & 8.31 \\
& Ours    & \underline{85.62} & \underline{65.12} & \textbf{0.91} & \textbf{0.95} & \textbf{0.98} \\
\hline
\multirow{2}{*}{TCGA-LUAD}
& Vanilla & \textbf{98.40} & \textbf{98.42} & 3.31 & 6.85 & 11.38 \\
& Ours    & \underline{97.96} & \underline{97.90} & \textbf{1.56} & \textbf{1.72} & \textbf{5.85} \\
\hline
\end{tabular}%
}
\end{minipage}%
\hfill
\begin{minipage}[c]{0.45\textwidth}
\centering
\vspace{3mm}
\includegraphics[width=\textwidth]{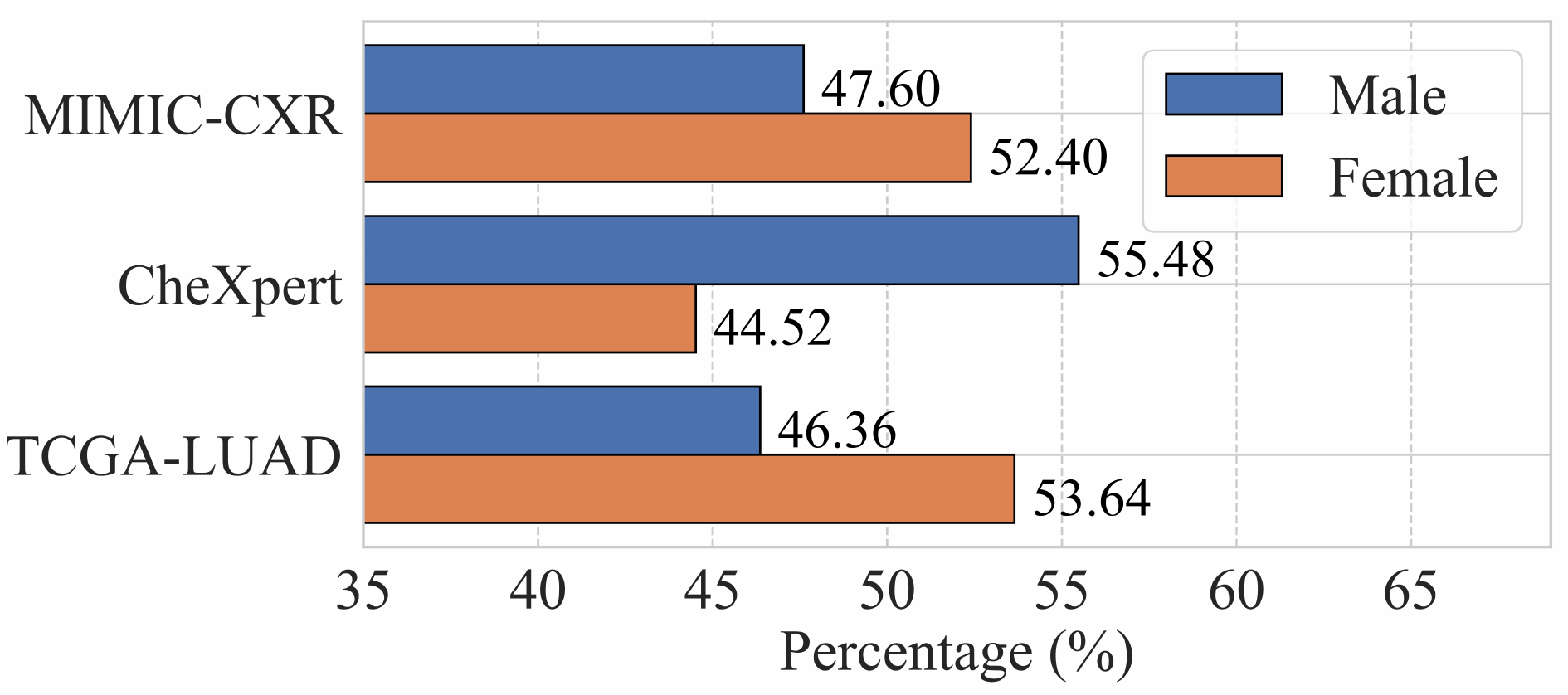}
\caption{Gender distribution across three datasets.}
\label{fig:gender_distribution}
\end{minipage}
\end{figure*}

To assess robustness across different sensitive attributes, we evaluate our method using age (binarized at 60). As shown in Table~\ref{tab:age_comparison}, our method consistently improves fairness on both attributes across all datasets. On MIMIC-CXR, for instance, age-based fairness gains are consistent (EOp: 4.72 $\to$ 1.56, EOd: 9.38 $\to$ 1.45). Overall, these results indicate that our method is insensitive to the choice of sensitive attribute and remains effective under different demographic imbalances.
\subsubsection{Parameter Ablation Study}
We further examine the effect of hyperparameters on model performance and fairness.

\begin{figure*}[t!]
\centering
\begin{minipage}[c]{0.48\textwidth}
\centering
\captionof{table}{Ablation studies on different noise strength $\eta$.}
\label{tab:eta}
\vspace{9pt}
\resizebox{\textwidth}{!}{
\begin{tabular}{c|c|cc|ccc}
\toprule
\textbf{Dataset} & $\eta$ & \textbf{$\text{ACC}_{\%}$} & \textbf{$\text{AUC}_{\%}$} & \textbf{$\text{EOp}_{e-2}$} & \textbf{$\text{EOd}_{e-3}$} &
\textbf{$\text{DP}_{e-2}$}\\ \hline
\multirow{5}{*}{MIMIC} 
& 0   & \textbf{85.46} & \textbf{63.85} & 5.10 & 10.59 & 6.31\\ 
& 0.1 & \underline{84.75} & \underline{62.90} & 3.48 & 3.98 & 3.75\\ 
& 0.2 & 84.26 & 62.10 & \textbf{2.02} & \textbf{2.25} & \textbf{3.53} \\  
& 0.3 & 81.00 & 60.27 & \underline{2.25} & \underline{2.33} & \underline{3.72} \\
& 0.4 & 80.70 & 59.95 & 2.67 & 2.59 & 4.50\\ \hline
\multirow{5}{*}{CheXpert} 
& 0   & \textbf{85.87} & \textbf{65.93} & 5.05 & 11.18 & 8.86 \\ 
& 0.1 & \underline{85.16} & \underline{64.50} & 4.90 & 7.82 & 4.83 \\ 
& 0.2 & 84.59 & 63.11 & \underline{3.47} & \underline{3.68} & \underline{2.64}\\  
& 0.3 & 84.02 & 63.00 & \textbf{2.52} & \textbf{3.24} & \textbf{1.02} \\
& 0.4 & 82.05 & 61.07 & 4.58 & 9.11 & 3.63 \\ \hline
\multirow{5}{*}{TCGA} 
& 0   & \textbf{98.47} & \textbf{98.58} & 3.57 & 7.21 & 11.94\\ 
& 0.1 & \underline{98.16} & \underline{98.43} & 2.38 & 5.14 & 9.09\\ 
& 0.2 & 97.85 & 98.20 & \textbf{1.39} & \textbf{1.46} & \textbf{6.89}\\  
& 0.3 & 97.16 & 98.09 & \underline{1.69} & \underline{2.82} & \underline{7.69}\\
& 0.4 & 96.55 & 97.63 & 1.78 & 2.98 & 7.95\\ \bottomrule
\end{tabular}}
\end{minipage}%
\hfill
\begin{minipage}[c]{0.50\textwidth}
\centering
\includegraphics[width=\textwidth]{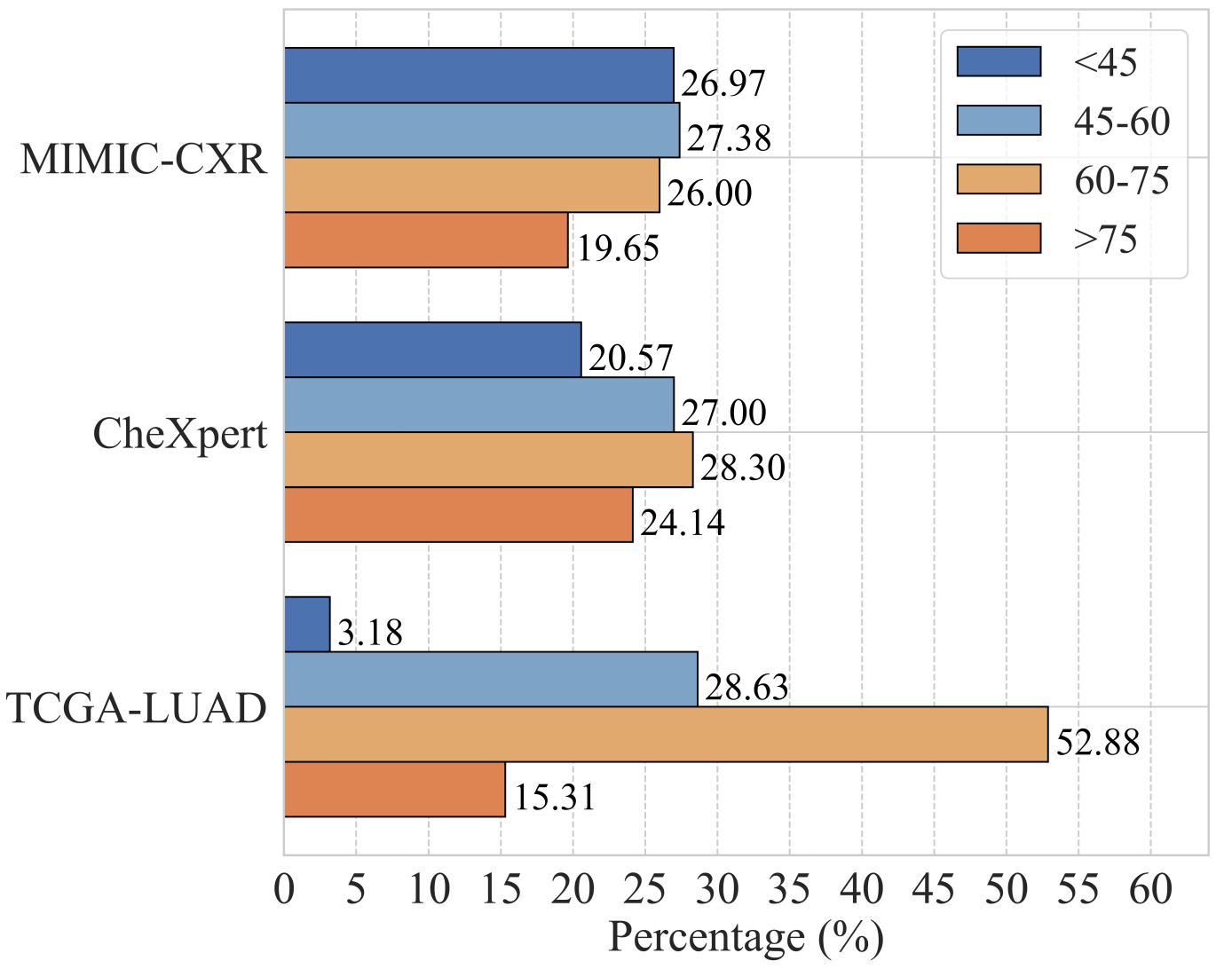}
\vspace{-6pt}
\caption{Age distribution across three datasets.}
\label{fig:age_distribution}
\end{minipage}
\end{figure*}

\textbf{Ablation on Noise Strength $\eta$:} Table \ref{tab:eta} shows the effect of varying the noise strength $\eta$ in the mask generation.
We observe that as $\eta$ increases, the performance gradually decreases. This decline suggests that larger $\eta$ values result in stronger masking, which filters out more information, potentially removing some relevant diagnostic features.
As for fairness, EOp, EOd, and DP metrics initially improve with a slight increase in $\eta$, indicating that a small amount of noise helps reduce bias. 
However, when $\eta$ becomes too large, fairness begins to deteriorate. This is likely because the generator struggles to optimize effectively under strong masking, thus impacting both performance and fairness. 
Based on these findings, we set $\eta = 0.2$ in the remaining experiments to balance performance and fairness.

\begin{table*}[t]
\centering
\begin{minipage}{0.48\textwidth}
\centering
\caption{Ablation studies on different $\alpha$.}
\resizebox{\textwidth}{!}{
\begin{tabular}{c|c|cc|ccc}
\toprule
\textbf{Dataset} & $\alpha$ & \textbf{$\text{ACC}_{\%}$} & \textbf{$\text{AUC}_{\%}$} & \textbf{$\text{EOp}_{e-2}$} & \textbf{$\text{EOd}_{e-3}$} & \textbf{$\text{DP}_{e-2}$} \\ \hline
\multirow{5}{*}{MIMIC} 
& 0 & 86.53 & 63.05 & 5.56 & 2.32 & 3.58 \\
& 1 & 86.46 & 62.79 & 4.50 & 1.62 & 2.54 \\
& 2 & \underline{86.58} & \underline{63.14} & 3.84 & 1.41 & 1.89 \\
& 3 & \textbf{86.92} & \textbf{64.68} & \underline{3.55} & \underline{1.20} & \underline{1.22} \\
& 4 & 85.99 & 63.02 & \textbf{3.51} & \textbf{1.13} & \textbf{1.02} \\
\hline
\multirow{5}{*}{CheXpert}
& 0 & 85.41 & 64.93 & 2.42 & 5.12 & 4.48 \\
& 1 & 85.21 & 65.24 & 1.33 & 3.36 & 3.41 \\
& 2 & \underline{85.52} & \underline{65.45} & 1.30 & 3.12 & 3.08 \\
& 3 & 85.29 & 65.18 & \underline{1.22} & \underline{2.26} & \underline{2.84} \\
& 4 & \textbf{85.60} & \textbf{65.52} & \textbf{1.00} & \textbf{1.52} & \textbf{1.67} \\
\hline
\multirow{5}{*}{TCGA} 
& 0 & \underline{98.42} & \underline{98.39} & 2.38 & 4.79 & 9.54 \\
& 1 & \textbf{98.70} & \textbf{98.51} & 1.89 & 3.42 & 7.74 \\
& 2 & 98.02 & 97.94 & 1.75 & 1.50 & 7.50 \\
& 3 & 98.55 & 98.25 & \underline{1.43} & \underline{1.47} & \underline{7.01} \\
& 4 & 98.19 & 98.01 & \textbf{1.39} & \textbf{1.46} & \textbf{6.89} \\
\bottomrule
\end{tabular}}
\label{tab:alpha}
\end{minipage}
\hfill
\begin{minipage}{0.48\textwidth}
\centering
\caption{Ablation studies on different $\beta$.}
\resizebox{\textwidth}{!}{
\begin{tabular}{c|c|cc|ccc}
\toprule
\textbf{Dataset} & $\beta$ & \textbf{$\text{ACC}_{\%}$} & \textbf{$\text{AUC}_{\%}$} & \textbf{$\text{EOp}_{e-2}$} & \textbf{$\text{EOd}_{e-3}$} & \textbf{$\text{DP}_{e-2}$} \\ \hline
\multirow{5}{*}{MIMIC} 
& 0 & 80.69 & 54.65 & \textbf{1.08} & \textbf{0.98} & \textbf{2.63} \\
& 1 & 84.98 & 62.34 & \underline{1.39} & \underline{1.87} & \underline{2.78} \\
& 2 & 85.03 & 62.87 & 2.03 & 2.55 & 3.34 \\
& 3 & \underline{86.98} & \underline{64.40} & 5.56 & 5.06 & 4.73 \\
& 4 & \textbf{87.79} & \textbf{64.80} & 7.58 & 5.50 & 5.31 \\
\hline
\multirow{5}{*}{CheXpert} 
& 0 & 71.15 & 59.75 & \textbf{0.54} & \textbf{1.31} & \textbf{0.42} \\
& 1 & 84.59 & 65.13 & \underline{3.32} & \underline{1.54} & \underline{2.02} \\
& 2 & 85.90 & 65.40 & 3.50 & 2.04 & 3.03 \\
& 3 & \underline{86.73} & \underline{65.89} & 3.75 & 2.45 & 3.36 \\
& 4 & \textbf{87.74} & \textbf{66.65} & 4.50 & 5.25 & 3.48 \\
\hline
\multirow{5}{*}{TCGA} 
& 0 & 72.70 & 92.01 & \textbf{1.03} & \textbf{1.42} & \textbf{6.45} \\
& 1 & {97.85} & {98.20} & \underline{1.39} & \underline{1.46} & \underline{6.89} \\
& 2 & 98.16 & 98.63 & 1.97 & 2.27 & 9.30 \\
& 3 & \underline{98.30} & \underline{98.71} & 2.03 & 3.55 & 9.35 \\
& 4 & \textbf{98.33} & \textbf{99.24} & 4.76 & 5.81 & 9.97 \\
\bottomrule
\end{tabular}}
\label{tab:beta}
\end{minipage}
\end{table*}
\textbf{Ablation on Entropy Loss Regularization $\alpha$:} Table \ref{tab:alpha} demonstrates the impact of varying the entropy loss regularization parameter $\alpha$.
As $\alpha$ increases, we observe a gradual decrease in performance, while fairness metrics improve.
This trade-off occurs because the entropy loss  $\mathcal{H}(\hat{S})$ encourages the generator to produce predictions that make the discriminator’s task of predicting $S$ more challenging, which promotes fairness. 
Based on this trade-off, we set $\alpha = 1$ in the remaining experiments to achieve an optimal balance between performance and fairness.

\textbf{Ablation on Loss Regularization $\beta$:} The results are presented in Table \ref{tab:beta}. This parameter controls the weight of the generator loss term $\mathcal{I}(\hat{Y}, Y)$, which is designed to preserve performance while applying masks to the input image. As $\beta$ increases, we observe a gradual increase in both accuracy and AUC, accompanied by a decrease in fairness metrics (EOp, EOd and DP). This trade-off occurs because a higher $\beta$ value allows the generator to focus more on preserving diagnostic performance, but at the cost of reduced fairness, as the generator becomes less constrained in limiting sensitive attribute influence.

\section{Conclusion}
\label{sec:conclusion}
In this work, we presented SPARC, a framework that makes Path-Specific Counterfactual Fairness practically achievable in high-dimensional settings. The core insight is that enforcing PSCF can be reduced to a causal conditional independence constraint, $\mathcal{I}(S; X | Y) = 0$, which replaces intractable counterfactual density estimation with a discriminative optimization objective. We proved that this conditional independence suffices to eliminate the direct causal effect, and that the optimization is directly observable through Equalized Odds, bridging causal fairness theory with standard evaluation metrics.
Empirically, SPARC consistently achieves superior fairness–accuracy trade-offs across multiple datasets, architectures and sensitive attributes. The explainability analysis further confirms that SPARC effectively disentangles prediction-relevant and sensitive-attribute-related regions in the input space.

While our main experiments focus on binary sensitive attributes for clarity and comparability with the PSCF literature, the underlying MI reduction does not inherently require this restriction (Appendix~\ref{app:multi}). Extending SPARC to richer attribute definitions, more complex causal structures, and multi-task and multi-modal settings remains an interesting direction for future work.

%
%
\bibliographystyle{splncs04}
\bibliography{main}

\clearpage
\appendix
\setcounter{theorem}{0}
\renewcommand{\thetheorem}{\arabic{theorem}}
\section{Theoretical Proof}
\label{appendix:the}
In this section, we present the theoretical derivability and observability of the SPARC framework.

\begin{theorem}[Derivability]\label{theorem:2}
For random variables \( Y \), \( S \), \( X \) and \( \hat{Y} \), the conditional mutual information $\mathcal{I}(S;X \mid Y) = 0$ is a sufficient condition of DE$(S)=0$.
\end{theorem}
\begin{proof}
According to Eq. \ref{eq:2}, $\text{DE}(S)$ can be written as:
\begin{align}
&P(\hat{Y}_{S \leftarrow s^{+}|\pi_d,\, S \leftarrow s^{-}|\pi_i})
- P(\hat{Y}_{S \leftarrow s^{-}})\nonumber\\[1mm]
=&\sum_{X,Y}P(\hat{Y}\mid X)\cdot P(X\mid Y,do(s^+|\pi_d))P(Y\mid do(s^-|{\pi_i})) \nonumber\\
    -& \sum_{X,Y}P(\hat{Y}\mid X) \cdot P(X\mid Y,do(s^-|{\pi_d}))P(Y\mid do(s^-|{\pi_i})),\nonumber\\
    =\;& \sum_{X}P(\hat{Y}\mid X)\big[\sum_{Y}P(X\mid Y,do(s^+|\pi_d))P(Y\mid do(s^-|\pi_i)) \nonumber\\
    -& \sum_{Y}P(X\mid Y,do(s^-|\pi_d))P(Y\mid do(s^-|\pi_i))\big],
    \label{eqq}
\end{align}

According to the definition of $\mathcal{I}(S;X \mid Y) = 0$, it satisfies ($S\perp X \mid Y$), therefore, for any $\Delta s\in\{s^+,s^-\}$,

According to the assumption of ignorability and consistency \citep{pearl2000models}:
\begin{align}
P(X \mid Y, {do}(\Delta s \mid \pi_d)) = P(X \mid Y, \Delta s)
\end{align}

\text{Since } $\mathcal{I}(S;X\mid Y)=0 \iff P(X\mid Y,\Delta s)=P(X\mid Y)$, \text{ we have}
$P(X \mid Y, {do}(\Delta s \mid \pi_d)) = P(X \mid Y)$. Applying this condition to the bracket in Eq. \ref{eqq} yields:
\begin{align}
&\sum_Y P(X\mid Y)P(Y\mid {do}(s^-\mid \pi_i)) \nonumber\\
-&\sum_Y P(X\mid Y)P(Y\mid {do}(s^-\mid \pi_i))=0.
\end{align}
Therefore, the condition $\mathcal{I}(S;X\mid Y)=0$ is sufficient to ensure $\text{DE}(S)=0$.
\qed
\end{proof}

{\color{black}
\textbf{Remark (Sufficiency vs.\ Necessity).}
We emphasize that $\mathcal{I}(S;X\mid Y)=0$ is a sufficient but not necessary condition for PSCF. SPARC constrains $X$ directly, which is stricter than PSCF ($\text{DE}(S)=0$): it removes all residual dependence between $X$ and $S$ given $Y$, whereas PSCF only requires the component propagating to $\hat{Y}$ through $f_h$ to vanish. The price of this stricter formulation is that SPARC may also suppress directions in $X$ that carry $S$-information but contribute negligibly to $\hat{Y}$, which would be admissible under PSCF. In exchange, the constraint becomes a discriminative objective that is tractable in high-dimensional $X$. Our synthetic-SCM analysis (Appendix~\ref{app:synthetic}) indicates this gap is small in practice ($|\text{DE}(S)| = 0.027$ vs.\ target $0$).}

The theorem \ref{theorem:1} concludes that if we meet the requirement of \(\mathcal{I}(S;X \mid Y) = 0 \), PSCF will be satisfied, where $\pi_d$ will be pruned, showcasing that the adversarial training between $\mathbf{g}$ and $\mathbf{d}_Y$ are theoretically reaching PSCF. Subsequently, we further prove that $\mathcal{I}(S;\hat{Y} \mid Y) = 0$ can be the evaluation metric for the objective of adversarial training, shedding light on the fairness observation.
\begin{theorem}[Observability]
\label{theorem:3}
For random variables \( Y \), \( S \), \( X \), and \( \hat{Y} \), $\mathcal{I}(S;X \mid Y) = 0$
is a sufficient condition for Equalized Odds.
\end{theorem}

\begin{proof}
Since we are discussing under the circumstances of Fig. \ref{fig:fig1}, i.e., the SCM assumes that \( X \) is a complete mediator between \( \hat{Y} \) and \( (Y, S) \). This assumption implies that any dependence between \( \hat{Y} \) and \( (Y, S) \) can be fully mediated by \( X \). Using the law of total probability \cite{grimmett2020probability}, we can express  $P(\hat{Y} \mid Y, S)$  as follows:
\begin{align}
P(\hat{Y} \mid Y, S) = &\int P(\hat{Y} \mid X, Y, S) P(X \mid Y, S) \, dX \nonumber\\
=&\int P(\hat{Y} \mid X) P(X \mid Y, S) \, dX \nonumber\\
=& \int P(\hat{Y} \mid X) P(X \mid Y) \, dX \nonumber\\
=& P(\hat{Y} \mid Y).
\end{align}
Thus, the equation holds ($S\perp \hat{Y} \mid Y$), which is equivalent to the information theory form of equalized odds: $\mathcal{I}(S;\hat{Y} \mid Y) = 0$.
\qed
\end{proof}

Theorem \ref{theorem:2} establishes the observability of the SPARC framework by demonstrating that path-specific counterfactual fairness can be reduced to a verifiable conditional independence constraint, $\mathcal{I}(S;X \mid Y) = 0$. By proving that eliminating the dependence between the sensitive attribute $S$ and the intermediate representations $X$ (given the label $Y$) is sufficient to ensure $\mathcal{I}(S;\hat{Y} \mid Y) = 0$, this theorem provides a rigorous theoretical foundation for SPARC, as it aligns the mitigation of unfair causal effects directly with measurable statistical fairness metrics, ensuring that the model's optimization process is both tractable and theoretically grounded.

\section{Experiment Settings}
\label{app:a}
\subsection{Baselines}
We compare with the following baselines: Vanilla \citep{he2016deep}, Adversarial Debiasing (AD) \citep{zhang2018mitigating}, Fairness-Aware Adversarial Perturbation (FAAP) \citep{wang2022fairness}, SUBG \citep{YoubiIdrissi2021SimpleDB} and Group-DRO \citep{Sagawa2019DistributionallyRN}. Vanilla uses traditional gradient descent to optimize; AD introduces an adversarial network that attempts to detect sensitive attributes of the primary model’s output; FAAP generates adversarial perturbations that incorporate fairness constraints; SUBG aims to achieve competitive worst-group-accuracy through simple data balancing; and Group-DRO focuses on distributionally robust neural networks to improve generalization for worst-case scenarios across groups.

\subsection{Datasets}
\textbf{MIMIC-CXR} \cite{johnson2019mimic}: The MIMIC-CXR dataset, part of the MIMIC initiative, includes over 370,000 chest X-rays from more than 60,000 patients, labeled with conditions like pneumonia, pleural effusion, and lung lesions. This anonymized dataset is valuable for machine learning research, supporting tasks such as automated disease detection and clinical decision support, especially in critical care environments.

\noindent
\textbf{CheXpert} \cite{irvin2019chexpert}: The CheXpert dataset, created by Stanford University, contains over 224,000 chest X-rays annotated for 14 radiological findings, including lung opacity and cardiomegaly. Known for its high-quality labels obtained through a robust NLP-based pipeline, CheXpert is widely used in AI research for disease classification and anomaly detection in medical imaging.

\noindent
\textbf{TCGA-LUAD} \cite{albertina2016cancer}: This dataset from The Cancer Genome Atlas (TCGA) project focuses on lung adenocarcinoma, providing gene expression, mutation profiles, methylation, and clinical data. It is instrumental for research in lung cancer biomarkers, diagnostics, and targeted therapies, supporting advancements in cancer genomics and precision medicine.

\subsection{Fairness Metrics}
We evaluate fairness using the following metrics:

\noindent\textbf{Equalized Opportunity ($\text{EOp}$)} \cite{hardt2016equality} is a notion of nondiscrimination with respect to a specified protected attribute, measuring the disparity in true positive rates across sensitive groups. Specifically, 
\begin{equation}
    \text{EOp} = |P(\hat{Y}=1|S=1,Y=1) - P(\hat{Y}=1|S=0,Y=1)|. 
\end{equation}
\noindent\textbf{Equalized Odds ($\text{EOd}$)} \citep{hardt2016equality} quantifies the independence between the predicted diagnosis $\hat{Y}$ and the sensitive attribute $S$, conditioned on the ground truth $Y$, where we measure in the information theory form \citep{ghassami2018fairness, ni2025kernel, zamani2025information}:
\begin{equation}
    \text{EOd} = \mathcal{I}(S;\hat{Y}\mid Y).
\end{equation}

\noindent\textbf{Demographic Parity (DP)} \cite{dwork2012fairness} measures the disparity in prediction rates between different sensitive groups (e.g., $S=0$ vs. $S=1$). Formally, 
\begin{equation}
    \text{DP} = |P(\hat{Y}=1|S=1) - P(\hat{Y}=1|S=0)|,
\end{equation}
where a smaller DP value indicates fewer disparities between sensitive groups in predictions.
$\text{EOd}$ captures how much uncertainty about $\hat{Y}$ can be reduced by knowing $S$, under the condition of $Y$.

\section{Implementation details}
\label{sec:appendix_section}
\subsection{Hyperparameter Settings} 

In our experiments, we explored several key hyperparameters to evaluate their impact on performance and fairness. Specifically, the noise strength $\eta$, controlling the intensity of adversarial perturbations generated by the model's generator, is set to 0.2 in experiments. The fairness weighting parameter $\alpha$ is set to 1.0, balancing the entropy-based fairness loss with task performance.  Loss regularization $\beta$ is set to 1.0 to prioritize classification accuracy. The learning rate for the generator, discriminator, and feature extractor was initialized at $1 \times 10^{-4}$, with step-based learning rate schedulers applied to ensure convergence. Training and testing datasets were split with a ratio of 90\% to 10\%, ensuring sufficient data for both robust evaluation and training. For standard deviation measurement, we repeat our experiments for five times for calculation.

\subsection{Dataset-Specific Tasks} 
Regarding dataset labels, for the MIMIC-CXR \cite{johnson2019mimic} and CheXpert \cite{irvin2019chexpert} dataset, our main task focused on pneumonia classification, utilizing gender as the sensitive attribute (mapped as male = 0, female = 1). For the TCGA \cite{albertina2016cancer} dataset, we focused on detecting pathological stages of lung cancer, specifically distinguishing between early and late stages. We preprocess the dataset to ensure a balanced representation of these stages (Stage I A, Stage I B, Stage II A, and Stage II B are mapped to the early stage; Stage III A, Stage III B, Stage III C, Stage IV A, and Stage IV B are mapped to late stage), enabling robust evaluation of both performance and fairness metrics in the classification task. Also, gender is utilized as a sensitive attribute.

\subsection{Environments} The models are trained offline using PyTorch \citep{paszke2019pytorch} and executed on a machine equipped with an AMD EPYC 7763 64-Core Processor CPU @ 4.00GHz and an NVIDIA RTX 6000 Ada Generation GPU, running the Ubuntu 22.04.3 LTS operating system. The experiments were conducted within a Conda environment and a Docker container to ensure reproducibility and ease of deployment. The Conda environment was managed using Miniconda \cite{miniconda} version 23.9.0 (Python 3.10.13), while the Docker container \cite{docker} was built on Docker version 24.0.5 with a base image of ``nvidia/cuda:12.4.0-cudnn8-devel-ubuntu22.04" to support GPU acceleration. We will provide our conda environment, Docker container, and code implementations upon publication.

\section{Experiments}
\subsection{Computational Efficiency}

Table~\ref{tab:flops} reports the computational overhead introduced by SPARC's additional components relative to the ResNet50 backbone.

\begin{table}[h]
\centering
\caption{Computational cost of each component.}
\label{tab:flops}
\small
\begin{tabular}{l|c}
\toprule
\textbf{Component} & \textbf{GFLOPs} \\
\midrule
Backbone (ResNet50) & 13.86 \\
Generator $\mathbf{g}$ & 1.50 \\
MI Estimator $f_\omega$ & 0.04 \\
Downstream head $f_h$ & 0.17 \\
\midrule
\textbf{Total overhead} & \textbf{+12.0\%} \\
\bottomrule
\end{tabular}
\end{table}

The additional components introduce only a 12.0\% increase in FLOPs over the backbone, confirming that SPARC remains computationally practical for high-dimensional medical imaging tasks.
{\color{black}
\subsection{Synthetic SCM Validation of Causal Effects}
\label{app:synthetic}
To directly verify that SPARC reduces the path-specific direct effect rather than merely improving observational metrics, we construct a synthetic SCM in which the ground-truth $\text{DE}(S)$ and $\text{IE}(S)$ are explicitly computable. Following \citep{chiappa2018pathspecific}, we define
\begin{align}
S &\sim \text{Bern}(0.5), \quad Y = \mathbb{I}[U_Y + 0.6\,S > 0], \nonumber\\
X &= U_X + \alpha_d S + \alpha_i Y, \quad U_X, U_Y \sim \mathcal{N}(0,1),
\end{align}
where $\alpha_d$ controls the direct path $S \to X$ and $\alpha_i$ the indirect path $S \to Y \to X$, with $\alpha_d = \alpha_i = 1$. Ground-truth $\text{DE}$ and $\text{IE}$ are computed following Eq.~\ref{eq:2} by intervening on $S$ along $\pi_d$ while holding $\pi_i$ fixed, and vice versa.

\begin{table}[h]
\centering
\caption{Synthetic SCM validation. Results averaged over 5 seeds. Ground-truth $|\text{IE}(S)| = 0.182$, target $|\text{DE}(S)| = 0$.}
\label{tab:synthetic}
\small
\begin{tabular}{l|cc}
\toprule
\textbf{Method} & $|\text{DE}(S)|$ ($\downarrow$) & $|\text{IE}(S)|$ (preserve) \\
\midrule
Vanilla & 0.284 & 0.182 \\
AD      & 0.038 & 0.061 \\
Ours    & \textbf{0.027} & \textbf{0.171} \\
\bottomrule
\end{tabular}
\end{table}

As shown in Table~\ref{tab:synthetic}, SPARC reduces $|\text{DE}(S)|$ from 0.284 (Vanilla) to 0.027 while preserving $|\text{IE}(S)|$ at 0.171 (close to the ground truth 0.182). In contrast, AD reduces $|\text{DE}(S)|$ to 0.038 but collapses $|\text{IE}(S)|$ to 0.061, suppressing legitimate indirect information. This confirms the path-specific behavior of SPARC and complements the high-dimensional experiments in the main text.

\subsection{Hyperparameter Generalization}

To verify that our hyperparameter selection does not overfit to the test set, we re-split each dataset into 80\%/10\%/10\% train/validation/test partitions and select hyperparameters solely based on validation performance. Table~\ref{tab:hpgen} shows that the selected hyperparameters transfer well to the test set, with minimal performance gaps across all datasets.

\begin{table}[h]
\centering
\caption{Test performance using hyperparameters selected on the validation set vs.\ the test set.}
\label{tab:hpgen}
\small
\begin{tabular}{l|l|cc|cc}
\toprule
\textbf{Dataset} & \textbf{Method} 
  & \multicolumn{2}{c}{\textbf{ACC (\%)}} 
  & \multicolumn{2}{c}{\textbf{EOd ($\mathbf{\times 10^{-3}}$)}} \\
\cmidrule(lr){3-6}
& & Val & Test & Val & Test \\
\midrule
\multirow{2}{*}{MIMIC-CXR}
  & Vanilla & 86.23 & 84.03 & 9.98  & 10.45 \\
  & Ours    & 84.58 & 83.95 & 2.02  & 2.34  \\
\midrule
\multirow{2}{*}{CheXpert}
  & Vanilla & 86.05 & 85.02 & 10.97 & 11.42 \\
  & Ours    & 85.69 & 84.68 & 1.22  & 1.49  \\
\midrule
\multirow{2}{*}{TCGA-LUAD}
  & Vanilla & 98.34 & 98.68 & 7.28  & 7.21  \\
  & Ours    & 97.64 & 98.07 & 1.48  & 1.66  \\
\bottomrule
\end{tabular}
\end{table}

Our method consistently outperforms the Vanilla baseline on both validation and test splits, confirming that the reported improvements are not an artifact of test-set hyperparameter tuning.

\subsection{Extension to Multi-Attribute Settings}
\label{app:multi}
The MI formulation in Theorem~\ref{theorem:1} does not require $S$ to be binary: $\mathcal{I}(S; X \mid Y)$ remains well-defined for categorical or continuous $S$, with the discriminator $\mathbf{d}_Y$ only needing an output head matching the cardinality of $S$. As a preliminary verification, we evaluate a multi-attribute setting on MIMIC-CXR using the joint attribute (gender $\times$ age), obtaining ACC $83.84\%$ and EOd $2.61\times10^{-3}$, supporting the extension. For richer causal graphs, the same reduction extends by conditioning the discriminator on the parents along fair paths, analogous to conditioning on $Y$ here.
}

\end{document}